\documentclass{article}

\PassOptionsToPackage{numbers, sort&compress}{natbib}

\usepackage[final]{neurips_2019}

\usepackage[utf8]{inputenc} % allow utf-8 input
\usepackage[T1]{fontenc}    % use 8-bit T1 fonts
\usepackage{hyperref}       % hyperlinks
\usepackage{url}            % simple URL typesetting
\usepackage{booktabs}       % professional-quality tables
\usepackage{amsfonts}       % blackboard math symbols
\usepackage{nicefrac}       % compact symbols for 1/2, etc.
\usepackage{microtype}      % microtypography
\usepackage{graphicx}

\usepackage{algorithm}
\usepackage[noend]{algpseudocode}
\usepackage{float}

\usepackage{wrapfig}
\usepackage{xcolor} % should be unnecessary for final version
\usepackage{bm}
\usepackage{amsmath}
\newcommand{\tp}{^{\mathrm{T}}}

\DeclareMathOperator*{\argmax}{arg\,max}

\DeclareMathOperator{\E}{E}
\DeclareMathOperator{\betapdf}{Beta}

\DeclareMathOperator{\bernoullipdf}{Bernoulli}

\DeclareMathOperator{\normalpdf}{N}

\newcommand{\mdp}{\mathcal{M}}

\newcommand{\beginsupplement}{%
	\setcounter{table}{0}
	\renewcommand{\thetable}{S\arabic{table}}%
	\setcounter{figure}{0}
	\renewcommand{\thefigure}{S\arabic{figure}}%
}

\title{
	Machine Teaching of Active Sequential Learners
}

\author{%
	Tomi Peltola\\
	\texttt{tomi.peltola@aalto.fi} \And
	Mustafa Mert \c{C}elikok\\
	\texttt{mustafa.celikok@aalto.fi}
	\And
	Pedram Daee\\
	\texttt{pedram.daee@aalto.fi}
	\And
	Samuel Kaski\\
	\texttt{samuel.kaski@aalto.fi}\\
	Helsinki Institute for Information Technology HIIT\\
	Department of Computer Science, %\\
	Aalto University, Helsinki, Finland
}

\begin{document}
	
	\maketitle
	
	\begin{abstract}
		Machine teaching addresses the problem of finding the best training data that can guide a learning algorithm to a target model with minimal effort. In conventional settings, a teacher provides data that are consistent with the true data distribution. However, for sequential learners which actively choose their queries, such as multi-armed bandits and active learners, the teacher can only provide responses to the learner’s queries, not design the full data. In this setting, consistent teachers can be sub-optimal for finite horizons. We formulate this sequential teaching problem, which current techniques in machine teaching do not address, as a Markov decision process, with the dynamics nesting a model of the learner and the actions being the teacher's responses. Furthermore, we address the complementary problem of learning from a teacher that plans: to recognise the teaching intent of the responses, the learner is endowed with a model of the teacher. We test the formulation with multi-armed bandit learners in simulated experiments and a user study. The results show that learning is improved by (i) planning teaching and (ii) the learner having a model of the teacher. The approach gives tools to taking into account strategic (planning) behaviour of users of interactive intelligent systems, such as recommendation engines, by considering them as boundedly optimal teachers.

	\end{abstract}
	
	\section{Introduction}
	Humans, casual users and domain experts alike, are increasingly interacting with artificial intelligence or machine learning based systems. As the number of interactions in human--computer and other types of agent--agent interaction is usually limited, these systems are often based on active sequential machine learning methods, such as multi-armed bandits, Bayesian optimization, or active learning. These methods explicitly optimise for the efficiency of the interaction from the system's perspective. On the other hand, for goal-oriented tasks, humans create mental models of the environment for planning their actions to achieve their goals \cite{Newell:1972:HPS:1095704,the_cog_sci_book}. In AI systems, recent research has shown that users form mental models of the AI's state and behaviour \cite{DBLP:journals/corr/ChandrasekaranY17,Williams:2019:AII:3290605.3300677}. Yet, the statistical models underlying the active sequential machine learning methods treat the human actions as passive data, rather than acknowledging the strategic thinking of the user. %of the system.
	
	Machine teaching studies a complementary problem to active learning: how to provide a machine learner with data to learn a target model with minimal effort \cite{goldman1995complexity,zhu2015machine,zhu2018overview}. Apart from its fundamental machine learning interest, machine teaching has been applied to domains such as education \cite{rafferty2016faster} and adversarial attacks \cite{mei2015using}. In this paper, we study the machine teaching problem of active sequential machine learners: the learner sequentially chooses queries and the teacher provides responses to them. Importantly, to steer the learner towards the teaching goal, the teacher needs to appreciate the order of the learner's queries and the effect of the responses on it. Current techniques in machine teaching do not address such interaction. Furthermore, by viewing users as boundedly optimal teachers, and solving the (inverse machine teaching) problem of how to learn from the teacher's responses, our approach provides a way to formulate models of strategically planning users in interactive AI systems.
	
	Our main contributions are (i) formulating the problem of machine teaching of active sequential learners as planning in a Markov decision process, (ii) formulating learning from the teacher's responses as probabilistic inverse reinforcement learning, (iii) implementing the approach in Bayesian Bernoulli multi-armed bandit learners with arm dependencies, and (iv) empirically studying the performance in simulated settings and a user study. Source code is available at \url{https://github.com/AaltoPML/machine-teaching-of-active-sequential-learners}.

	\section{Related work}
	
	Most work in machine teaching considers a batch setting, where the teacher designs a minimal dataset to make the learner learn the target model \cite{goldman1995complexity,zhu2015machine,zhu2018overview}. Some works have also studied sequential teaching, but in different settings from ours: Teaching methods have been developed to construct batches of state-action trajectories for inverse reinforcement learners \cite{cakmak2012algorithmic,brown2019machine}. Variations on teaching online learners, such as gradient descent algorithms, by providing them with a sequence of $(\bm{x}, y)$ data points have also been considered \cite{lessard2019optimal,liu2017iterative,liu2018towards}. Teaching in the context of education, with uncertainty about the learner's state, has been formulated as planning in partially-observable Markov decision processes \cite{rafferty2016faster,whitehill2017approximately}. A theoretical study of the teacher-aware learners was presented in \cite{zilles2011models,doliwa2014recursive} where the teacher and the learner are aware of their cooperation. Compared to our setting, in these works, the teacher is in control of designing all of the learning data (while possibly using interaction to probe the state of the learner) and is not allowed to be inconsistent with regard to the true data distribution. Apart from \cite{brown2019machine, zilles2011models, doliwa2014recursive}, they also do not consider teacher-aware learners. Machine teaching can also be used towards attacking learning systems \cite{mei2015using}, and adversarial attacks against multi-armed bandits have been developed, by poisoning historical data \cite{ma2018data} or modifying rewards online \cite{jun2018adversarial}. The goal, settings, and proposed methods differ from ours. Relatedly, our teaching approach for the case of a bandit learner can been seen as a form of reward shaping, which aims to make the environment more supportive of reinforcement learning by alleviating the temporal credit assignment problem \cite{potential_reward_shaping}.

	The proposed model of the interaction between a teacher and an active sequential learner is a probabilistic multi-agent model. It can be connected to the overarching framework of interactive partially observable Markov decision processes (I-POMDPs; see Supplementary Section A for more details) \cite{gmytrasiewicz2005framework} and other related multi-agent models \cite{pynadath2002communicative,cooperative_irl,OliehoekAmato16book,albrecht2018autonomous}. I-POMDPs provide, in a principled decision-theoretic framework, a general approach to define multi-agent models that have recursive beliefs about other agents. This also forms a rich basis for computational models of theory of mind, which is the ability to attribute mental states, such as beliefs and desires, to oneself and other agents and is essential for efficient social collaboration \cite{baker2017rational,pedagogical_irl}. Our teaching problem nests a model of a teacher-unaware learner, forming a learner--teacher model. Teaching-aware learning adds a further layer, forming a nested learner--teacher--learner model, where the higher level learner models a teacher modelling a teaching-unaware learner. 
	Learning from humans with recursive reasoning was opined in \cite{ipomdp_humans}. To our knowledge, our work is the first to propose a multi-agent recursive reasoning model in the practically important case of multi-armed bandits, allowing us to learn online from the scarce data emerging from human--computer interaction.

	User modelling in human--computer interaction aims at improving the usability and usefulness of collaborative human--computer systems and providing personalised user experiences \cite{fischer2001user}. Machine learning based interactive systems extend user modelling to encompass statistical models interpreting user's actions. For example, in information exploration and discovery, the system needs to iteratively recommend items to the user and update the recommendations based on the user feedback \cite{Marchionini2006,ruotsalo2015interactive}. The current underlying statistical models use the user's response to the system's queries, such as \textit{did you like this movie?}, as data for building a relevance profile of the user. Recent works have investigated more advanced user models \cite{schmit18a,Daee2018overfitting}; however, as far as we know, no previous work has proposed statistical user models that incorporate a model of the user's mental model of the system.
	
	Finally, our approach can be grounded to computational rationality, which models human behaviour and decision making under uncertainty as expected utility maximisation, subject to computational constraints \cite{gershman2015computational}. Our model assumes that the teacher chooses actions proportional to their likelihood to maximise, for a limited horizon, the future accumulated utility.

	\section{Model and computation}
	
	We consider machine teaching of an active sequential learner, with the iterations consisting of the learner querying an input point $\bm{x}$ and the teacher providing a response $y$. First, the teaching problem is formulated as a Markov decision process, the solution of which provides a teaching policy. Then, learning from the responses provided by the teacher is formulated as an inverse reinforcement learning problem. We formulate the approach for general learners, and give a detailed implementation for the specific case of a Bayesian Bernoulli multi-armed bandit learner, which models arm dependencies.
	
	\subsection{Active sequential learning}
	
	Before considering machine teaching, we first define the type of active sequential learners considered. This also provides a baseline to which the teacher's performance is compared. The general definition encompasses multiple popular sequential learning approaches, including Bayesian optimisation and multi-armed bandits, which aim to learn fast, with few queries.
	
	An active sequential learner is defined by (i) a machine learning model relating the response $y$ to the inputs $\bm{x}$ through a function $f$, $y = f_{\bm{\theta}}(\bm{x})$, parameterised by $\bm{\theta}$, or through a conditional distribution $p(y \mid \bm{x}, \bm{\theta})$, (ii) a deterministic learning algorithm, fitting the parameters $\bm{\theta}$ or their posterior $p(\bm{\theta} \mid \mathcal{D})$ given a dataset $\mathcal{D} = \{(\bm{x}_1, y_1), \ldots, (\bm{x}_t, y_t)\}$, (iii) a query function that, possibly stochastically, chooses an input point $\bm{x}$ to query for a response $y$, usually formulated as utility maximisation.
	
	The dynamics of the learning process then, for $t = 1,\ldots,T$, consists of iterating the following steps:
	\begin{enumerate}
		\item Use the query function to choose a query $\bm{x}_t$.
		\item Obtain the response $y_t$ for the query $\bm{x}_t$ from a teacher (or some other information source).
		\item Update the training set to $\mathcal{D}_t =  \mathcal{D}_{t-1} \cup \{(\bm{x}_t, y_t)\} $ and the model correspondingly.
	\end{enumerate}
	The data produced by the dynamics forms a sequence, or \textit{history}, $h_T = \bm{x}_1, y_1, \bm{x}_2, y_2, \ldots, \bm{x}_T$ (we define the history to end at the input $\bm{x}_T$, before $y_T$, for notational convenience in the following).
	
	\paragraph{Bayesian Bernoulli multi-armed bandit learner} As our main application in this paper, we consider Bayesian Bernoulli bandits. At each iteration $t$, the learner chooses an arm $i_t \in \{1,\ldots,K\}$ and receives a stochastic reward $y_t \in \{0,1\}$, depending on the chosen arm. The goal of the learner is to maximise the expected accumulated reward $R_T = \E[\sum_{t=1}^T y_t]$. This presents an exploration--exploitation problem, as the learner needs to learn which arms produce reward with high probability.
	
	The learner associates each arm $k$ with a feature vector $\bm{x}_k \in \mathbb{R}^M$ and models the rewards as Bernoulli-distributed binary random variables
	\begin{equation}
	\begin{split}
	p_\mathcal{B}(y_t \mid \mu_{i_t}) & = \bernoullipdf(y_t \mid \mu_{i_t}) \label{eqn:basic_user_model}
	\end{split}   
	\end{equation}
	with reward probabilities $\mu_k  = \sigma(\bm{x}_k\tp \bm{\theta}), k=1,\ldots,K,$ where $\bm{\theta} \in \mathbb{R}^M$ is a weight vector and $\sigma(\cdot)$ the logistic sigmoid function. The linearity assumption could be relaxed, for example, by encoding the $\bm{x}_k$'s using suitable basis functions or Gaussian processes. The Bayesian learner has a prior distribution on the model parameters, here assumed to be a multivariate normal, $\bm{\theta} \sim \normalpdf(\bm{0}, \tau^2 \bm{\mathrm{I}})$, with mean zero and diagonal covariance matrix $\tau^2 \bm{\mathrm{I}}$. Given a collected set of arm selections and reward observations at step $t$, $\mathcal{D}_t = \{(i_1, y_1),\ldots,(i_t, y_t)\}$ (or equivalently $\mathcal{D}_t = (h_t, y_t)$), the posterior distribution of $\bm{\theta}$, $p(\bm{\theta} \mid \mathcal{D}_t)$ is computed.
	
	The learner uses a bandit arm selection strategy to select the next arm to query about. Here, we use Thompson sampling \cite{thompson1933likelihood}, a practical and empirically and theoretically well-performing algorithm \cite{russo2018tutorial}; other methods could easily be used instead. The next arm is sampled with probabilities proportional to the arm maximising the expected reward, estimated over the current posterior distribution:
	\begin{equation}
	\Pr(i_{t+1}\!=\!k)\!=\!\!\int\!\!I(\argmax_j \mu_j\!=\!k \mid \bm{\theta}) p(\bm{\theta} \mid \mathcal{D}_t) d\bm{\theta},\label{eqn:ts}
	\end{equation}%
	where $I$ is the indicator function. This can be realised by first sampling a weight vector $\bm{\theta}$ from $p(\bm{\theta} \mid \mathcal{D}_t)$, computing the corresponding $\bm{\mu}^{(\bm{\theta})}$, and choosing the arm with the maximal reward probability, $i_{t+1} = \argmax_k \mu^{(\bm{\theta})}_k$.

	\begin{figure*}
		\centering
		\includegraphics[scale=0.66]{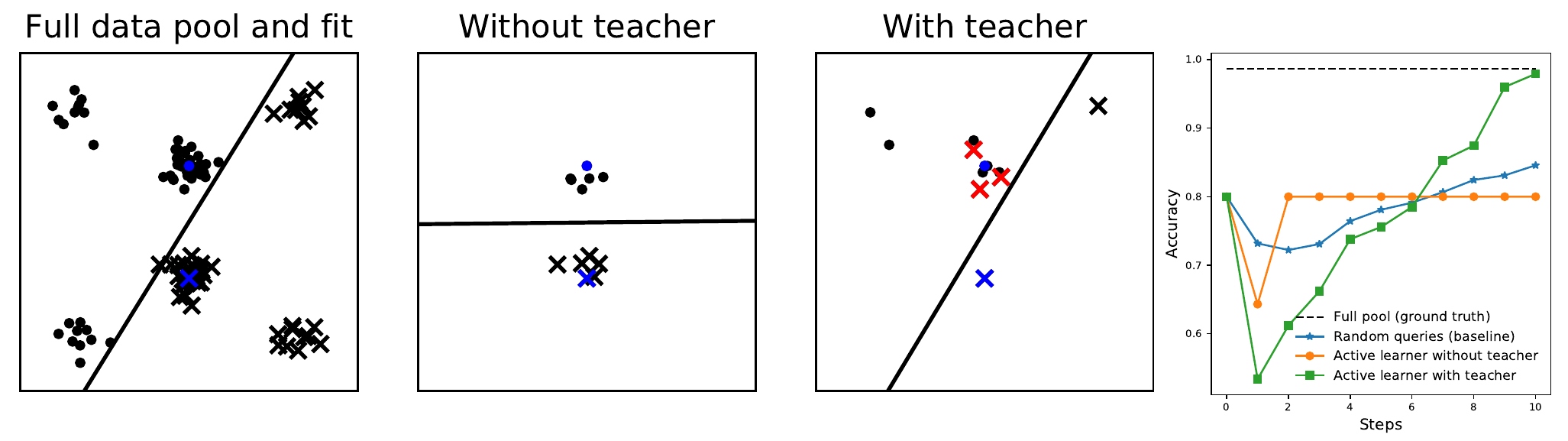}
		\caption{
			Example of teaching effect on pool-based logistic regression active learner. Using uncertainty sampling for queries, the learner fails to sample useful points from the pool in 10 iterations to learn a good decision boundary ("Without teacher"; starting from blue training data). A planning teacher can help the learner sample more representative points by switching some labels ("With teacher"; switched labels are shown in red). The average accuracy improvement is shown in the right panel. Details of the setting are given in Supplementary Section B.
		}\label{fig:active_learning_example}
	\end{figure*}
	
	\begin{wrapfigure}{r}{0.5\textwidth}
		\vspace*{-5mm}\includegraphics[scale=0.65]{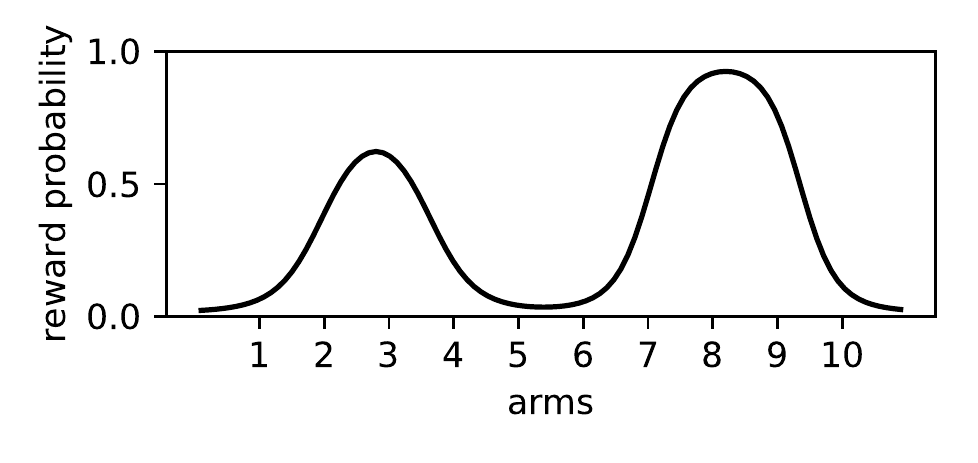}
		\centering
		\caption{
			Example of the teaching effect on a multi-armed bandit learner. With the environmental reward probabilities shown in the figure, consider the first query being arm 6. The reward probability for the arm is low, so $y_1 = 0$ with high probability for a naive teacher. Yet, the optimal action for a planning teacher is $y_1=1$, because the teacher can anticipate that this will lead to a higher probability for the learner to sample the next arm near the higher peak. Details on the setting are given in Supplementary Section C.
		}
		\label{fig:mab_teaching_example}
		\vspace*{-5mm}
	\end{wrapfigure}
	
	\subsection{Machine teaching of active sequential learner}
	
	In standard active sequential learning, the responses $y_t$ are assumed to be generated by a stationary data-generating mechanism as independent and identically distributed samples. We call such a mechanism a \textit{naive teacher}. Our machine teaching formulation replaces it with a \textit{planning teacher} which, by choosing $y_t$ carefully, aims to steer the learner towards a teaching goal with minimal effort.
	
	We formulate the teaching problem as a Markov decision process (MDP), where the transition dynamics follow from the dynamics of the sequential learner and the responses $y_t$ are the actions. The teaching MDP is defined by the tuple $\mdp = (\mathcal{H}, \mathcal{Y}, \mathcal{T}, \mathcal{R}, \gamma)$, where states $h_t \in \mathcal{H}$ correspond to the history, actions are the responses $y_t \in \mathcal{Y}$, transition probabilities $p(h_{t+1} \mid h_{t}, y_{t}) \in \mathcal{T}$ are defined by the learner's sequential dynamics, rewards $R_t(h_t) \in \mathcal{R}$ are used to define the teacher's goal, and $\gamma \in (0,1]$ is a discount factor (optional if $T$ is finite). The objective of the teacher is to choose actions $y_t$ to maximise the cumulative reward, called value, $V^{\pi}(h_1) = \E^{\pi}[\sum_{t=1}^T \gamma^{t-1} R_{t}(h_t)]$, where $T$ is the teacher's planning horizon and the expectation is over the possible stochasticity in the learner's queries and the teacher's policy. The teacher's policy $\pi(h_t, y_t) = p(y_t \mid h_t, \pi)$ maps the state $h_t$ to probabilities over the action space $\mathcal{Y}$. The solution to the teaching problem corresponds to finding the optimal teaching policy $\pi^*$.
	
	The reward function $R_t(h_t)$ defines the goal of the teacher. In designing a teaching MDP, as in reinforcement learning, its choice is crucial. In machine teaching, a natural assumption is that the reward function is parameterized by an optimal model parameter $\bm{\theta}^*$, or some other ground truth, known to the teacher but not the learner. For teaching of a supervised learning algorithm, the reward $R_t(h_t; \bm{\theta}^*)$ can, for example, be defined based on the distance of the learner's estimate of $\bm{\theta}$ to $\bm{\theta}^*$ or by evaluation of learner's predictions against the teacher's privileged knowledge of outcomes (Figure~\ref{fig:active_learning_example}).
	
	In the multi-armed bandit application, it is assumed that the teacher knows the true parameter $\bm{\theta}^*$ of the underlying environmental reward distribution and aims to teach the learner such that the accumulated environmental reward is maximised (Figure~\ref{fig:mab_teaching_example}). We define the teacher's reward function as $R_t(h_t; \bm{\theta}^*) = \bm{x}_t\tp \bm{\theta}^*$ (leaving out $\sigma(\cdot)$ to simplify the formulas for the teacher model).
	
	\paragraph{Properties of the teaching MDP}
	
	In Supplementary Section D, we briefly discuss the transition dynamics and state definition of the teaching MDP, and contrast it to Bayes-adaptive MDPs to better understand its properties. Finding the optimal teaching policy presents similar challenges to planning in Bayes-adaptive MDPs. Methods such as Monte Carlo tree search \cite{guez2013scalable} have been found to provide effective approaches.

	\subsection{Learning from teacher's responses}
	
	We next describe how the learner can interpret the teacher's responses, acknowledging the teaching intent. Having formulated the teaching as an MDP, the teacher-aware learning follows naturally as inverse reinforcement learning \cite{ramachandran2007bayesian,choi2011map}. We formulate a probabilistic teacher model to make the learning more robust towards suboptimal teaching and to allow using the teacher model as a block in probabilistic modelling.
	
	At each iteration $t$, the learner assumes that the teacher chooses the action $y_t$ with probability proportional to the action being optimal in value:
	\begin{equation}
	p_{\mdp}(y_t \mid h_{t}, \bm{\theta}^*) = \frac{\exp\left(\beta Q^*(h_{t}, y_t; \bm{\theta}^*) \right)}{\sum_{y'\in\mathcal{Y}} \exp\left(\beta Q^*(h_{t}, y'; \bm{\theta}^*) \right)},\label{eqn:la_user_model}   
	\end{equation}
	where $Q^*(h_{t}, y_t; \bm{\theta}^*)$ is the optimal state-action value function of the teaching MDP for the action $y_t$ (that is, the value of taking action $y_t$ at $t$ and following an optimal policy afterwards). Here $\beta$ is a \textit{teacher optimality parameter} (or inverse temperature; for $\beta=0$, the distribution of $y_t$ is uniform; for $\beta \rightarrow \infty$, the action with the highest value is chosen deterministically). From the teaching-aware learner's perspective, the teacher's $\bm{\theta}^*$ is unknown, and Equation~\ref{eqn:la_user_model} functions as the likelihood for learning about $\bm{\theta}$ from the observed teaching. In the bandit case, this replaces Equation~\ref{eqn:basic_user_model}. Note that the teaching MDP dynamics still follow from the teaching-unaware learner.
	
	\paragraph{One-step planning} Since our main motivating application is modelling users as boundedly optimal teachers, implemented for a Bernoulli multi-armed bandit system, it is interesting to consider the special case of one-step planning horizon, $T = 1$. The state-action value function $Q^*(h_t, y_t; \bm{\theta}^*)$ then simplifies to the rewards at the next possible arms, and the action observation model to
	\begin{equation}
	p_{\mdp}(y_t \mid h_t, \bm{\theta}^*) \propto \exp(\beta ((\bm{\theta}^*)\tp \bm{X}\tp \bm{p}_{h_t, y_t})),\label{eqn:la_user_model_1step}
	\end{equation}
	where $\bm{p}_{h_t, y_t} = [p_{1,h_t, y_t}, \ldots, p_{K,h_t, y_t}]\tp$ collects the probabilities of the next arm given action $y_t \in \{0,1\}$ at the current arm $\bm{x}_t$ in $h_t$, as estimated according to the teaching MDP, and $\bm{X} \in \mathbb{R}^{K \times M}$ collects the arm features into a matrix. Note that the reward of the current arm does not appear in the action probability\footnote{It cancels out. The teacher cannot affect the arm choice anymore, as it has already been made.}. For deterministic bandit arm selection strategies, the transition probabilities $p_{k, h_t, y_t}$ for each of the two actions would have a single $1$ and $K-1$ zeroes (essentially picking one of the possible arms), giving the action probability an interpretation as a preference for one of the possible next arms. For stochastic selection strategies, such as Thompson sampling, the interpretation is similar, but the two arms are now weighted averages, $\bar{\bm{x}}_{y_t=0} = \bm{X}\tp \bm{p}_{h_t, y_t=0}$ and $\bar{\bm{x}}_{y_t=1} = \bm{X}\tp \bm{p}_{h_t, y_t=1}$. An algorithmic overview of learning with a one-step planning teacher model is given in Supplementary Section E.
	
	For an illustrative example, consider a case with two independent arms ($\bm{x}_1 = [1,0]$ and $\bm{x}_2 = [0,1]$), with the first arm having a larger reward probability than the other ($\theta_1^ *> \theta_2^*$). The optimal teaching action is then to give $y_t=1$ for queries on arm 1 and $y_t=0$ for arm 2. A teaching-unaware learner will still need to query both arms multiple times to identify the better arm. A teaching-aware learner (when $\beta \rightarrow \infty$) can identify the better arm from a single query (on either arm), since the likelihood function tends to the step function $I(\theta_1^* > \theta_2^*)$. This demonstrates that the teaching-aware learner can use a query to reduce uncertainty about other arms even in the extreme case of independent arms.
	
	\paragraph{Incorporating uncertainty about the teacher}
	
	Teachers can exhibit different kinds of strategies. To make the learner's model of the teacher robust to different types of teachers, we formulate a mixture model over a set of alternative strategies. Here, for the multi-armed bandit case, we consider a combination of a teacher that just passes on the environmental reward (naive teacher, Equation~\ref{eqn:basic_user_model}) and the planning teacher (Equation~\ref{eqn:la_user_model}):
	\begin{equation}
	\begin{aligned}
	p_{\mathcal{B} / \mdp}(y_t \mid h_t, \bm{\theta}^*, \alpha) = &(1 - \alpha) p_\mathcal{B}(y_t \mid \mu_{i_t}) + \alpha p_{\mdp}(y_t \mid h_t, \bm{\theta}^*),\label{eqn:mix_user_model}
	\end{aligned}
	\end{equation}
	where $\alpha \in (0,1)$ is a mixing weight and $\mu_{i_t} = \sigma(\bm{x}_{i_t}\tp \bm{\theta}^*)$ is the reward probability of the latest arm in the history $h_t$. A beta prior distribution, $\alpha \sim \betapdf(1, 1)$, is assumed for the mixing weight.
	
	\subsection{Computational details for Bayesian Bernoulli multi-armed bandits}
	
	Computation presents three challenges: (i) computing the analytically intractable posterior distribution of the model parameters $p(\bm{\theta} \mid \mathcal{D}_t)$ or $p(\bm{\theta}^*,\alpha \mid \mathcal{D}_t)$, (ii) solving the state-value functions $Q^*$ for the teaching MDP, and (iii) computing the Thompson sampling probabilities that are needed for the state-value functions.
	
	We implemented the models in the probabilistic programming language Pyro (version 0.3, under PyTorch v1.0) \cite{bingham2018pyro} and approximate the posterior distributions with Laplace approximations \citep[Section~4.1]{Gelman2013}. In brief, the posterior is approximated as a multivariate Gaussian, with the mean defined by the maximum a posteriori (MAP) estimate and the covariance matrix being the negative of the inverse Hessian matrix at the MAP estimate. In the mixture model, the mixture coefficient $\alpha \in (0,1)$ is transformed to the real axis via the logit function before computing the approximation.
	
	The inference requires computing the gradient of the logarithm of the unnormalised posterior probability. For the teacher model, this entails computing the gradient of the logarithm of Equation~\ref{eqn:la_user_model} at any value of the model parameters, which requires solving and computing the gradients of the optimal state-action value functions $Q^*$ with respect to $\bm{\theta}^*$. To solve the $Q^*$ for both of the possible observable actions $y_t = 0$ and $y_t = 1$, we compute all the possible trajectories in the MDP until the horizon $T$ and choose the ones giving maximal expected cumulative reward. \citet{choi2011map} show that the gradients of $Q^*$ exist almost everywhere, and that the direct computation gives a subgradient at the boundaries where the gradient does not exist.
	
	We mainly focus on one-step planning ($T=1$) in the experiments. For long planning horizons and stochastic arm selection strategies, the number of possible trajectories grows too fast for the exact exhaustive computation to be feasible ($K^{T}$ trajectories for each initial action). In our multi-step experiments, we approximate the forward simulation of the MDP with \textit{virtual arms}: instead of considering all possible next arms given an action $y_t$ and weighting them with their selection probabilities $\bm{p}_{h_t, y_t}$, we update the model with a virtual arm that is the selection-probability-weighted average of the next possible arms $\bar{\bm{x}}_{h_t, y_t} = \bm{X}\tp \bm{p}_{h_t, y_t}$ (for deterministic strategies, this is exact computation). The virtual arms do not correspond to real arms in the system but are expectations over the next arms. This leads to $2^{T-1}$ trajectories to simulate for each initial action. Moreover, for any trajectory of actions $y_{1},\ldots,y_{T}$, this approximation gives $Q(h_1, y_1; \bm{\theta}^*) \approx (\bm{\theta}^*)\tp \bm{X}\tp \sum_{t=1}^{T} \gamma^{t-1} \bm{p}_{h_t, y_t}$ and if we cache the sum of the discounted transition probabilities for each trajectory from the forward simulation, we can easily find the optimal $Q^*$ at any value of $\bm{\theta}^*$ as required for the inference.
	
	Computing the next arm probabilities for the $Q^*$ values requires computing the actual Thompson sampling probabilities in Equation~\ref{eqn:ts} instead of just sampling from it. As the sigmoid function is monotonic, one can equivalently compute the probabilities as $\Pr(i_{t+1} = k) = \int I(\argmax_j z_j = k) p(\bm{z} \mid \mathcal{D}_t) d\bm{z}$ where $\bm{z} = \bm{X} \bm{\theta}^*$. As $p(\bm{\theta}^* \mid \mathcal{D}_t) \approx \normalpdf(\bm{\theta}^* \mid \bm{m}, \bm{\Sigma})$, $\bm{z}$ has multivariate normal distribution with mean $\bm{X} \bm{m}$ and covariance $\bm{X} \bm{\Sigma} \bm{X}\tp$. The selection probabilities can then be estimated with Monte Carlo sampling. We further use Rao-Blackwellized estimates $\Pr(i_{t+1} = k) \approx \frac{1}{L} \sum_{l=1}^L \Pr(z_k > \max_{j \neq k} z_j \mid \bm{z}_{-k}^{(l)})$, with $L$ Monte Carlo samples drawn for $\bm{z}_{-k}$ ($\bm{z}$ with $k$th component removed) and $\Pr(z_k > \max_{j \neq k} z_j \mid \bm{z}_{-k}^{(l)})$ being the conditional normal probability of component $z_k$ being larger than the largest component in $\bm{z}_{-k}$. 
	
	\section{Experiments}
	
	We perform simulation experiments for the Bayesian Bernoulli multi-armed bandit learner, based on a real dataset, to study (i) whether a teacher can efficiently steer the learner towards a target to increase learning performance, (ii) whether the ability of the learner to recognise the teaching intent increases the performance, (iii) whether the mixture model is robust to assumptions about the teacher's strategy, and (iv) whether planning multiple steps ahead improves teaching performance. We then present results from a proof-of-concept study with humans. Supplementary Section F.1 includes an additional experiment studying the teaching of an uncertainty-sampling-based logistic regression active learner, showing that teaching can improve learning performance markedly.

	\subsection{Simulation experiments}

	We use a word relevance dataset for simulating an information retrieval task. In this task, the user is trying to teach a relevance profile to the learner in order to reach her target word. The Word dataset is a random selection of 10,000 words from Google's Word2Vec vectors, pre-trained on Google News dataset \cite{mikolov2013distributed}. We reduce the dimensionality of the word embeddings from the original 300 to 10 using PCA. Feature vectors are mean-centred and normalised to unit length. We report results,  with similar conclusions, on two other datasets in Supplementary Section F.2.
	
	We randomly generate 100 replicate experiments: a set of 100 arms is sampled without replacement and one arm is randomly chosen as the target $\hat{\bm{x}}\in \mathbb{R}^{M}$. The ground-truth relevance profile is generated by first setting $\hat{\bm{\theta}}^* = [c, d\hat{\bm{x}}] \in \mathbb{R}^{M+1}$, where $c=-4$ is a weight for an intercept term (a constant element of $1$ is added to the $\bm{x}$s) and $d=8$ is a scaling factor. Then, the ground-truth reward probabilities are computed as $\hat{\mu}_k = \sigma(\bm{x}_k\tp\hat{\bm{\theta}}^*)$ for each arm $k$ (Supplementary Figure S2 shows the mean reward probability profile). To reduce experimental variance for method comparison, we choose one of the arms randomly as the initial query for all methods.

	\begin{wraptable}{r}{6.0cm}
		\vspace*{-7mm}
		\caption{Teacher--learner pairs.}\label{tbl:models}
		\begin{tabular}{rccc}
			\toprule
			& \multicolumn{3}{c}{\textbf{Learner's model of teacher}} \\
			\cmidrule(lr){2-4}
			\textbf{Teacher} & naive & planning & mixture \\
			\cmidrule(lr){1-1}
			\cmidrule(lr){2-4}
			naive & N-N & N-P & N-M\\
			planning & P-N & P-P & P-M\\
			\bottomrule
		\end{tabular}
		\vspace*{-3mm}
	\end{wraptable}
	
	We compare the learning performances of different pairs of simulated teachers and learners  (Table~\ref{tbl:models}). A naive teacher (N), which does not intentionally teach, passes on a stochastic binary reward (Equation~\ref{eqn:basic_user_model}) based on the ground truth $\hat{\mu}_k$ as its action for arm $k$ (the standard bandit assumption). A planning teacher (P) uses the probabilistic teaching MDP model (Equation~\ref{eqn:la_user_model_1step} for one-step and Equation~\ref{eqn:la_user_model} for multi-step) based on the ground truth $\hat{\bm{\theta}}^*$ to plan its action. We use $\hat{\beta} = 20$ as the planning teacher's optimality parameter and also set $\beta$ of the learner's teacher model to the same value. For multi-step models, we set $\gamma_{t} = \frac{1}{T}$, so that they plan to maximise the average return up to horizon $T$. The learners are named based on their models of the teacher: a teaching-unaware learner learns based on the naive teacher model (N; Equation~\ref{eqn:basic_user_model}) and teaching-aware learner models the planning teacher (P; Equation~\ref{eqn:la_user_model_1step} or Equation~\ref{eqn:la_user_model}). Mixture model (M) refers to the learner with a mixture of the two teacher models  (Equation~\ref{eqn:mix_user_model}).
	
	Expected cumulative reward and concordance index are used as performance measures (higher is better for both). Expected cumulative reward measures how efficiently the system can find high reward arms and is a standard bandit benchmark value. Concordance index is equivalent to the area under the receiver operating characteristic curve. It is a common performance measure for information retrieval tasks. It estimates the probability that a random pair of arms is ordered in the same order by their ground truth relevances and the model's estimated relevances; 0.5 corresponds to random and 1.0 to perfect performance.

	\begin{figure*}[t]
		\centering
		\includegraphics{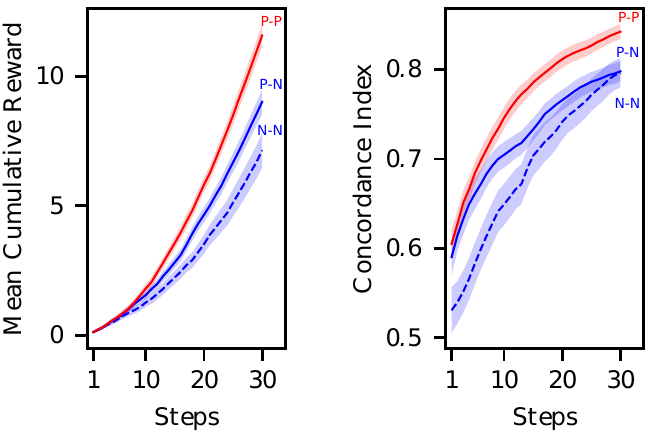} $\;\;\;\;\;$
		\includegraphics{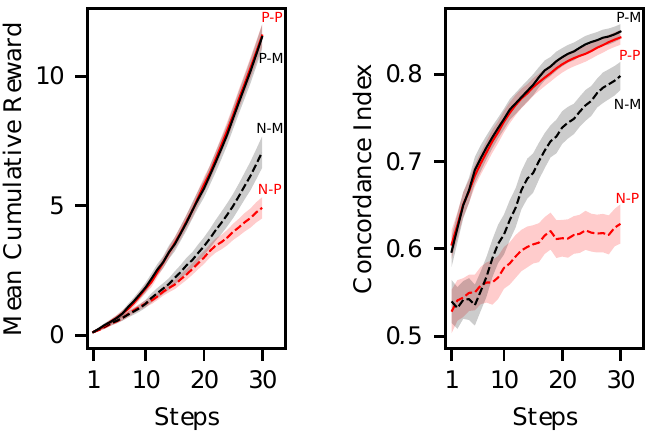}
		\caption{\textbf{Left-side panels:} Planning teacher improves performance, both when the learner's teacher model is naive (P-N) or planning (P-P), over naive teacher (N-N). \textbf{Right-side panels:} Naive teacher with a learner expecting a planning teacher (N-P) degrades performance. Learners with the mixture teacher model attain similar performance to matched models (P-M vs P-P and N-M vs N-N (left)). Lines show the mean over 100 replications and shaded area the 95\% confidence intervals for the mean. See Table~\ref{tbl:models} for key to the abbreviations.}
		\label{fig:sim_results}
	\end{figure*}
	
	\subsection{Simulation results}

	\begin{wrapfigure}{r}{0.5\textwidth}
		\centering
		\vspace*{-14mm}\includegraphics[scale = 0.95]{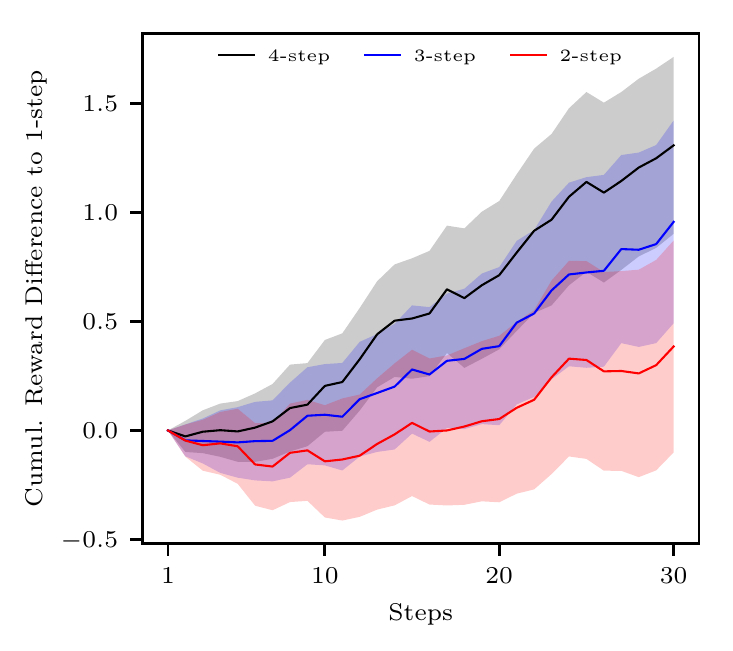}
		\caption{Teachers planning for multiple steps ahead improve over 1-step (P-P) in performance.% Lines show the mean over 50 replications and shaded area the 95\% confidence intervals for the mean.
		}\label{fig:multi_step_results}
		%\vspace*{-15mm}
		%\end{wrapfigure}
		
		\vspace*{2mm}
		
		%\begin{wrapfigure}{r}{0.5\textwidth}
		%\centering
		\includegraphics{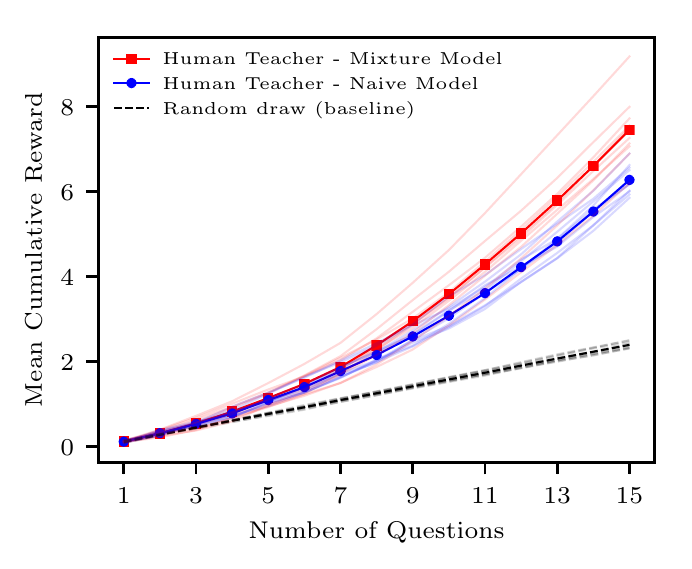}
		\caption{
			The accumulated reward was consistently higher for the participants when interacting with a learner having the mixture teacher model, compared to a learner with the naive teacher model. Shaded lines show the mean performance (over the 20 target words) of individual participants. Solid lines show the mean over the participants. Random arm sampling is shown as baseline.
		}\label{fig:user_study_cum_reward_all}
		\vspace*{-20mm}
	\end{wrapfigure}

	\paragraph{Teaching improves performance}

	Figure~\ref{fig:sim_results} shows the performance of different combinations of pairs of teachers and learners (where planning teachers have planning horizon $T=1$). The planning teacher can steer a teacher-unaware learner to achieve a marked increase in performance compared to a naive teacher (P-N vs N-N; left-side panels), showing that intentional teaching makes the reward signal more supportive of learning. The performance increases markedly further when the learner models the planning teacher (P-P; left-side panels). The improvements are seen in both performance measures, and the concordance index implies particularly that the proposed model learns faster about relevant arms and also achieves higher overall performance at the end of the 30 steps.

	\paragraph{Mixture model increases robustness to assumptions about the teacher}
	
	A mismatch of a naive teacher with a learner expecting a planning teacher (N-P) is markedly detrimental to performance (Figure~\ref{fig:sim_results} right-side panels). The mixture model guards against the mismatch and attains a performance similar to the matching assumptions (P-M vs P-P and N-M vs N-N).
	
	\paragraph{Planning for multiple steps increases performance further}
	
	Figure~\ref{fig:multi_step_results} shows the cumulative reward difference for matching planning teacher--learner pairs (P-P) when planning two to four steps ahead compared to one step. There is a marked improvement especially when going to 3-step or 4-step planning horizon. 
	
	\paragraph{Sensitivity analysis} Sensitivity of the results to the simulated teacher's optimality parameter $\hat{\beta}$ (performance degrades markedly for small values of $\hat{\beta}$) and to the number of arms (500 instead of 100; results remain qualitatively similar) are shown in Supplementary Section F.2.

	\subsection{User experiment}

	We conducted a proof-of-concept user study for the task introduced above, using a subset of 20 words % (from the Word dataset)
	on ten university students and researchers. The goal of the study was introduced to the participants as helping a system to find a target word, as fast as possible, by providing binary answers (yes/no) to the system's questions: ``Is this word relevant to the target?'' A target word was given to the participants at the beginning of each round (for twenty rounds; each word chosen once as the target word). Details of the study setting are provided in Supplementary Section G.

	Participants achieved noticeably higher average cumulative reward when interacting with a learner having the mixture teacher model, compared to a learner with the naive teacher model (Figure~\ref{fig:user_study_cum_reward_all}, red vs blue). This difference was at a significant level (p-value $<$ 0.01) after 12 questions, computed using paired sample t-test (see Supplementary Section G for p-values per step). 
	
	\section{Discussion and conclusions}
	
	We introduced a new sequential machine teaching problem, where the learner actively chooses queries and the teacher provides responses to them. This encompasses teaching popular sequential learners, such as active learners and multi-armed bandits. The teaching problem was formulated as a Markov decision process, the solution of which provides the optimal teaching policy. We then formulate teacher-aware learning from the teacher's responses as probabilistic inverse reinforcement learning. Experiments on Bayesian Bernoulli multi-armed bandits and logistic regression active learners demonstrated improved performance from teaching and from learning with teacher awareness. Better theoretical understanding of the setting and studying a more varied set of assumptions and approaches to planning for both the teacher and the teacher-aware learner are important future directions.
	
	Our formulation provides a way to model users with strategic behaviour as boundedly optimal teachers in interactive intelligent systems. We conducted a proof-of-concept user study, showing encouraging results, where the user was tasked to steer a bandit system towards a target word. To scale the approach to more realistic systems, for example, to interactive exploratory information retrieval \cite{exploratory_search_new}, of which our user study is a simplified instance, or to human-in-the-loop Bayesian optimisation \cite{brochu2010bayesian}, where the user might not possess the exact knowledge of the goal, future work should consider incorporating more advanced cognitive models of users. As an efficient teacher (user) needs to be able to model the learner (system), our results also highlight the role of understandability and predictability of interactive systems for the user as an important design factor, not only for user experience, but also for the statistical modelling in the system.
	
	While we focused here on teachers with bounded, short-horizon planning (as we would not expect human users to be able to predict behaviour of interactive systems for long horizons), scaling the computation to larger problems is of interest. Given the similarity of the teaching MDP to Bayes-adaptive MDPs (and partially observable MDPs), planning methods developed for them could be used for efficient search for teaching actions. The teaching setting has some advantages here: as the teacher is assumed to have privileged information, such as a target model, that information could be used to generate a reasonable initial policy for choosing actions $y$. Such policy could be then refined, for example, using Monte Carlo tree search. The teacher-aware learning problem is more challenging, as inverse reinforcement learning requires handling the planning problem in an inner loop. Considering the application and adaptation of state-of-the-art inverse reinforcement learning methods for teacher-aware learning is future work.
	
	\subsubsection*{Acknowledgments}
	
	This work was financially supported by the Academy of Finland (Flagship programme: Finnish Center for Artificial Intelligence, FCAI; grants 319264, 313195, 305780, 292334). Mustafa Mert Çelikok is partially funded by the Finnish Science Foundation for Technology and Economics KAUTE. We acknowledge the computational resources provided by the Aalto Science-IT Project. We thank Antti Oulasvirta and Marta Soare for comments that improved the article.
	
	%\bibliography{neurips_references}
	\bibliographystyle{unsrtnat}

	\newpage
	%###########################################
	%###########################################
	%###########################################
	%###########################################
	%###########################################
	\appendix
	
	\beginsupplement
	
	\section{Connection to I-POMDPs and multi-agent opponent modelling}
	
	Recursive modelling of the opponents' reasoning is studied in multi-agent and game theory communities. These methods employ theory-of-mind-like models to reason about the opponents' behaviour.
	
	Interactive POMDPs are a general recursive model of multi-agent interaction where the state space of a POMDP is extended by adding possible models of the opponents \citep{gmytrasiewicz2005framework2}. An I-POMDP agent maintains a belief over the original states of the POMDP and the possible models of its opponents. Opponent models are nested in the sense that a level-$k$ I-POMDP has level-$k-1$ opponent models in its state space. Level $0$ is a POMDP with no opponent models, where the effects of the actions of others are subsumed into transition dynamics.
	
	I-POMDPs suffer from three curses: (1) curse of dimensionality; because the state space is over the states and joint beliefs, (2) curse of history; because the policy space grows exponentially with respect to the planning horizon, and (3) curse of nestedness. The curse of nestedness is due to the fact that the solution of each level depends recursively on the solutions of lower-level I-POMDPs. Even though there are approximate particle filtering approaches, previous work has shown that they scale poorly even to medium-sized problems \cite{Ng2010TowardsAI}. So far, most of the recent work in I-POMDPs has been evaluated only on toy domains \cite{han2018learning}.  
	
	Our proposed learner with the planning teacher model is a level-1 agent who models its teacher as a level-0 agent. The level-0 agent is a teaching MDP who subsumes the query selection behaviour of the learner into the transition dynamics $\mathcal{T}^+$. Different from general I-POMDPs, we do not have any environment states. The teacher and learner are interacting directly. The level-0 agent models the learner as the environment, and the level-1 learner models the teacher as a level-0 agent.
	The learner's goal is to learn the function behind the teacher's actions (represented as the reward function) and its teacher model changes the learning rule via the likelihood. 
	
	The opponent model space of I-POMDPs usually contains an additional type of models called sub-intentional models. These are simple models such as an opponent who acts uniformly at random, or one that chooses its actions from a fixed yet unknown distribution. Our mixture model can be seen as a level-1 learner which maintains a belief over two possible opponent models: level-0 planning teacher model and the sub-intentional naive teacher model. This belief is represented by the posterior of the mixture coefficient $\alpha$.  
	
	In summary, our modelling can be seen as part of the I-POMDP framework, yet we differ in terms of objectives, modes of interaction, and environment settings. These differences allow us to have reasonable improvements in terms of computational and sample complexity.

	%\clearpage
	
	\section{Details for the example of teaching effect on pool-based logistic regression active learner}
	
	Figure 1 in the main text shows an example of teaching effect on a pool-based logistic regression active learner. The learner is a logistic regression model with L2 regularization. It is initialised with 2 data points (one for each label from the large clusters in the middle) and has a pool of 60 unlabeled data points for which it can query the label. The generated dataset follows a pattern where uncertainty sampling is known to fail \cite{yang2018benchmark}. Ten iterations are run in the example.
	
	The learner uses uncertainty sampling for selecting queries: the next query is chosen as the data point $\bm{x}$, for the label of which the current logistic regression model has the largest entropy, $-\sum_{y \in \{0,1\}} p_{\bm{\theta}}(y \mid x) \log p_{\bm{\theta}}(y \mid x)$. After obtaining a label, the model is updated and a new query made. Each unlabeled data point can be queried only once.
	
	The teacher plans for the full horizon of 10 iterations, with the reward defined at the terminal state of the horizon as the accuracy of the classification. The teacher has knowledge of the labels for the full pool of data, so the accuracy for the reward is evaluated using the full pool.
	
	\section{Details for the example of the teaching effect on a multi-armed bandit learner}
	
	Figure 2 in the main text shows an example of the teaching effect on a multi-armed bandit learner. The example follows the Bayesian Bernoulli multi-armed bandit setting. The ten arms are located evenly spaced on the x-axis from $0$ to $1$ (in the main text, the figure's x-axis labels show the arm numbers). A three-dimensional feature space is constructed with the first feature being constant $1$, second based on an RBF kernel at $0.2$ on the x-axis, and the third an RBF kernel at $0.8$ on the x-axis. The reward function is linear in this feature space, with weight $\bm{\theta}^* = [-4, 4.5, 6.5]$.
	
	Consider the first query, before the learner has any observations, being arm $6$. The reward probability for the arm $6$ is low ($0.06$), so $y_1 = 0$ with high probability when there is no teacher. Yet, the optimal action for a one-step planning teacher (i.e., teacher's horizon $T=1$) is $y_1=1$, because the teacher can anticipate this leading to a higher probability of sampling the next arm near the higher peak.
	
	\section{Properties of the teaching MDP and comparison to Bayes-adaptive Markov decision processes}
	
	We briefly discuss the transition dynamics and state definition of the teaching MDP, and contrast it to Bayes-adaptive MDPs (BAMDPs) to better understand its properties. The teaching MDP is defined in Section 3.2 of the main text.
	
	The transition probabilities $p(h_{t+1} \mid h_{t}, y_{t})$ can be decomposed into two factors: (1) an update of the learner's model given the new observation $y_t$ at the query point $\bm{x}_t$ (corresponding to step 3 in the learning dynamics), which contributes a deterministic factor to the transition (equivalent to adding $y_t$ to the history), and (2) the selection of a new query point $\bm{x}_{t+1}$ using the query function (corresponding to step 1 in the learning dynamics). If the query function is stochastic, this contributes a stochastic factor. Otherwise, the transitions are deterministic. 
	
	Depending on the learner model, instead of defining the state as the full history of the process, it is possible to define the state as a combination of the latest query point $\bm{x}_t$ and the model parameters (or posterior parameters), if the model parameters form a sufficient statistic for the history with respect to the model and the query function.
	
	The structure of the teaching MDP is similar in two respects to Bayes-adaptive MDPs (and partially observable MDPs), which describe an agent's uncertainty about the underlying transition dynamics (or state) \cite{duff2002optimal, ghavamzadeh2015bayesian}: (1) the state definition includes the full history of the process (or its sufficient statistic), and (2) the transition probabilities can be decomposed into the two steps, featuring an update of the model and sampling of a transition conditional on the model. The main difference is that the teaching MDP does not describe the agent's uncertainty about the process, but the dynamics themselves evolve according to the two steps. In particular, the teacher is assumed to know the next arm probabilities (so there is no uncertainty about the dynamics, although the dynamics can be stochastic, so the teacher does not know the exact next arm). Moreover, the model updated in BAMDP is directly a model of the transition dynamics (with observations being of form $(s, a, s')$, that is, new state $s'$ given previous state $s$ and action $a$), whereas in the teaching MDP, it is the learner's model (with observations of form $(\bm{x}, y)$, response $y$ given a query point $\bm{x}$). The next arm probabilities then follow from the definition of the query algorithm, and, for Bayesian bandit learner, can be computed as expectations over the learner's posterior distribution. The teaching MDP could be naturally extended to a teaching BAMDP, if the teacher has uncertainty about the learner (for example, which query strategy the learner uses). In any case, similar challenges to BAMDPs are faced in finding the optimal policy. Planning methods, such as Monte Carlo tree search \cite{guez2013scalable2}, have been found to provide effective approaches.
	
	\subsection{Further BAMDP background and using them to model multi-armed bandit problems} 
	
	A Bayes-adaptive Markov decision process (BAMDP) extends the state space of an MDP with unknown transition dynamics $\mathcal{T}$ by adding posterior beliefs about the transition dynamics \cite{duff2002optimal}. This new state is also called the information state. The information state of a BAMDP at a given time is $s^{+} = (s, h)$ where $s$ is the original MDP state and $h$ is the history of transitions observed so far. Then, a belief over transitions $P(\mathcal{T} \mid h)$ is maintained. Often in practice, this is a parametric distribution and the information state maintains the sufficient statistics instead of the whole history. The transition dynamics of the BAMDP in the extended state space is $\mathcal{T}^+((s,h), a, (s',h')) = \mathbb{E}_{P(\mathcal{T} \mid h)}[\mathcal{T}(s,a,s')]$. BAMDP's actions and rewards are the same as the original MDP. Solving this BAMDP with the $\mathcal{T}^+$ yields the Bayes-optimal solution for the original MDP, balancing the exploration with exploitation optimally. 
	
	A multi-armed bandit problem can be expressed as a BAMDP where information states contain only the histories since there are no environment states. In that case, the history $h$ will consist of played arms and observed rewards so far. Let $h'=har$, where $h'$ is the played arm $a$ and observed reward $r$ appended to the history $h$. Then the transitions are defined as $\mathcal{T}^+(h,a,h') = \mathbb{E}_{P(\mathcal{R}\mid h)}[P(r|a)]$ where $P(\mathcal{R}\mid h)$ is the posterior belief about reward probabilities of all arms, given history $h$. The solution to this BAMDP, with a given prior $P(\mathcal{R})$ as the starting state, provides the Bayes-optimal solution to the multi-armed bandit problem.
	
	One can derive popular MAB algorithms within the BAMDP as well. For instance, Thompson sampling procedure at each time step can be seen as estimating $\mathcal{T}^+$ with a single sample from $P(\mathcal{R}\mid h)$ instead of the full expectation, and then acting with a greedy argmax policy on it.
	
	Different from the BAMDP formulations of multi-armed bandits, where the action space is choosing among the arms, the teaching MDP takes the perspective of the reward generating mechanism (teacher). While the uncertainty for a BAMDP agent is about the reward distribution, for a teacher, the uncertainty is about which arm the learner will sample next.

	\section{Algorithmic Overview for Bandit Learner}
	
	Algorithm~\ref{alg:bandit} describes a bandit learner, with naive, planning, or mixture model of the teacher. The learner nests a model of the teacher (Algorithm~\ref{alg:user}). Note that the history $h_t$, up to $t$, is defined as the sequence $h_t = \bm{x}_1, y_1, \bm{x}_2, y_2, \ldots, \bm{x}_t$, which ends at the input $\bm{x}_t$ and doesn't include $y_t$. We also assume $p(\bm{\theta} \mid h_0, y_0) = p(\bm{\theta})$ in Algorithm~\ref{alg:bandit}. The naive, planning, and mixture likelihoods are defined in Equations 1, 4, and 5 in the main text.
	
	\begin{algorithm}[H]
		\caption{Bandit Learner}
		\label{alg:bandit}
		\begin{algorithmic}
			\For{$t \gets 1$ to $T$}
			\State $i_t \gets \text{thompson\_sample}(p(\bm{\theta} \mid h_{t-1}, y_{t-1}), \bm{X})$
			\Comment{select the next arm}
			\State $h_t \gets h_{t-1},y_{t-1},\bm{x}_{i_t}$
			\Comment{update history}
			\State $y_t \gets \text{teacher}(i_t)$ 
			\Comment{get the response from the teacher}
			\If{learners\_teacher\_model $=$ naive}
			\State $\mathcal{L} \gets p_\mathcal{B}(y_t \mid \bm{x}_{i_t}, \bm{\theta})$ \Comment{naive teacher likelihood}
			\EndIf 
			\If{learners\_teacher\_model $=$ planning}
			\State $\mathcal{L} \gets  \text{planning\_teacher\_model}(h_{t}, y_t)$ \Comment{planning teacher likelihood}
			\EndIf 
			\If{learners\_teacher\_model $=$ mixture}
			\State $\mathcal{L} \gets (1 - \alpha) p_\mathcal{B}(y_t \mid \bm{x}_{i_t}, \bm{\theta}) +  \alpha \text{planning\_teacher\_model}(h_t, y_t)$ \Comment{mixture likelihood}
			\EndIf 
			\State $p(\bm{\theta}, \alpha \mid y_t,  h_t) \gets \text{posterior\_update}(\mathcal{L}, p(\bm{\theta}, \alpha \mid h_{t-1}, y_{t-1}))$ 
			\Comment{($\alpha$ if mixture model)}
			\EndFor
		\end{algorithmic}
	\end{algorithm}

	\begin{algorithm}[H]
		\caption{One-step Planning Teacher Model}
		\label{alg:user}
		\begin{algorithmic}
			\Function{planning\_teacher\_model}{$h,y$}
			\ForAll{$y' \in \{0,1\}$}
			\State $p_{n}(\bm{\theta} \mid h,y') \gets \text{naive\_update}(h, y')$
			\Comment{simulate posterior update of naive learner}
			\State $\bm{p}_{h,y'} \gets \text{estimate\_thompson\_probabilities}(p_{n}(\bm{\theta} \mid h,y'), \bm{X})$
			\Comment{next arm probabilities}
			\EndFor
			\State \Return $p_{\mdp}(y \mid h, \bm{\theta}) \propto \exp(\beta (\bm{\theta}\tp\bm{X}\tp\bm{p}_{h,y}))$ \Comment{return the planning likelihood}
			\EndFunction
		\end{algorithmic}
	\end{algorithm}
	
	\section{Supplementary results for simulated experiments}
	
	\subsection{Logistic regression active learner}
	
	While our main experiments focus on the multi-armed bandit setting, we provide here an additional experiment for teaching of a logistic regression active learner. This experiment uses the Wine Quality dataset \cite{cortez2009modeling}, consisting of 4,898 instances of white wines with 11 continuous features and a ordinal output variable denoting wine quality. We transform the problem into a classification task by thresholding the quality.
	
	The learner is a logistic regression model with L2 regularization. It is initialised with 2 data points (one for each label from the large clusters in the middle) and has a further pool of 2000 unlabeled data points for which it can query the label. The learner uses uncertainty sampling for selecting queries: the next query is chosen as the data point $\bm{x}$ for the label of which the current logistic regression model has the largest entropy, $-\sum_{y \in \{0,1\}} p_{\bm{\theta}}(y \mid x) \log p_{\bm{\theta}}(y \mid x)$. After obtaining a label, the model is updated and a new query done. Each unlabeled data point can be queried only once. The rest of the dataset is used as a test dataset for evaluating the performance, with classification accuracy as the performance metric.
	
	The teacher plans for 1 step ahead, with full knowledge of the labels for the pool of 2000 data points (but not the test data). The reward is defined as the accuracy of the classifier.
	
	We run the experiment for a horizon of 100 steps, with 100 repetitions of randomly dividing the data into the training pool and test set.
	
\begin{figure*}[h!]
		\includegraphics[scale=0.8]{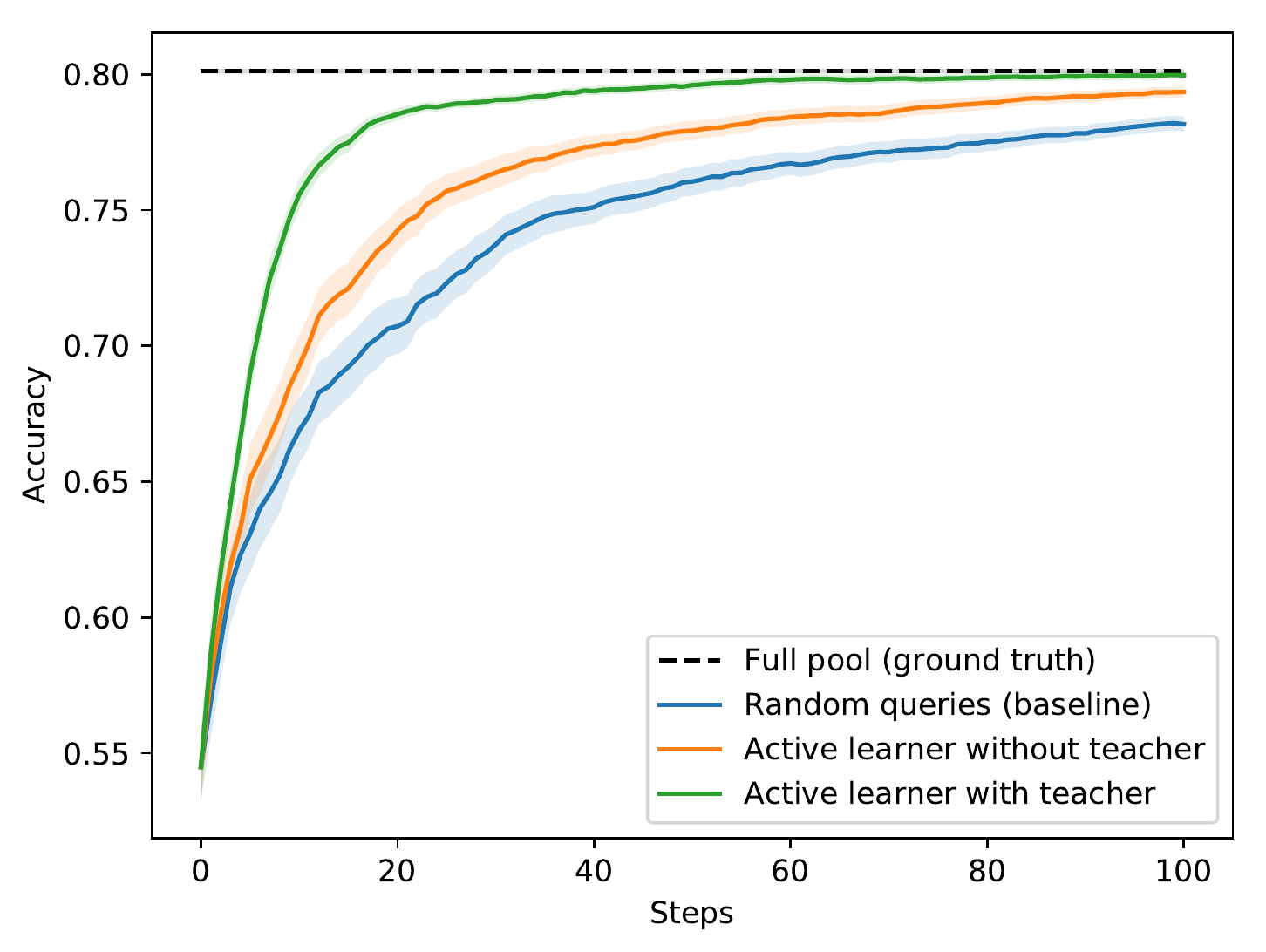}
		\centering
		\caption{Effect of teaching of pool-based logistic regression active learner in the Wine dataset. Lines show the mean over 100 replications and shaded area the 95\% confidence intervals for the mean.}
		\label{fig:active_learning_wine}
	\end{figure*}

	Figure~\ref{fig:active_learning_wine} compares the accuracy on the test set for the active learning with teacher (green), active learner without teacher (orange), and learner using random queries to gather more data (blue). The accuracy of the logistic regression model fitted to the full pool is shown for reference (black). The teacher improves the learning performance markedly, attaining performance close to the full pool model with around 20 training samples.
	
	\subsection{Multi-armed bandits}
	
	We provide further results for the simulation studies here, as listed below. The two further datasets are the following, corresponding roughly to data that would occur in tasks for recommendation and image search, respectively. The Wine Quality dataset \cite{cortez2009modeling} consists of 4,898 instances of white wines with 11 continuous features (and the ordinal output variable denoting wine quality which is not used here). The Leaf dataset \cite{leaf} consists of 340 instances with 14 features representing the shape and texture features of leaves from different plant species. For all datasets, all feature vectors are mean-centred and normalised to unit length. 
	
	Here, for the Word dataset, we also present a sensitivity analysis for a different ground-truth reward profile generated by $c=-2$ and $d=6$, referred to as the supplementary setting. The qualitative difference between the new profile and the profile from the main text is that in the former there are more arms with high reward probabilities. This makes the rewards more informative and supportive of learning. The figures related to the supplementary setting indicate that a more informative reward means smaller gains from a planning teacher steering a naive learner. However, the combination of a planning teacher and a teacher-aware learner still provides a considerable improvement.
	
	All simulation experiments were run on Linux computers, running PyTorch 1.0 and Pyro 0.3.
	
\begin{figure*}[h!]
		\includegraphics{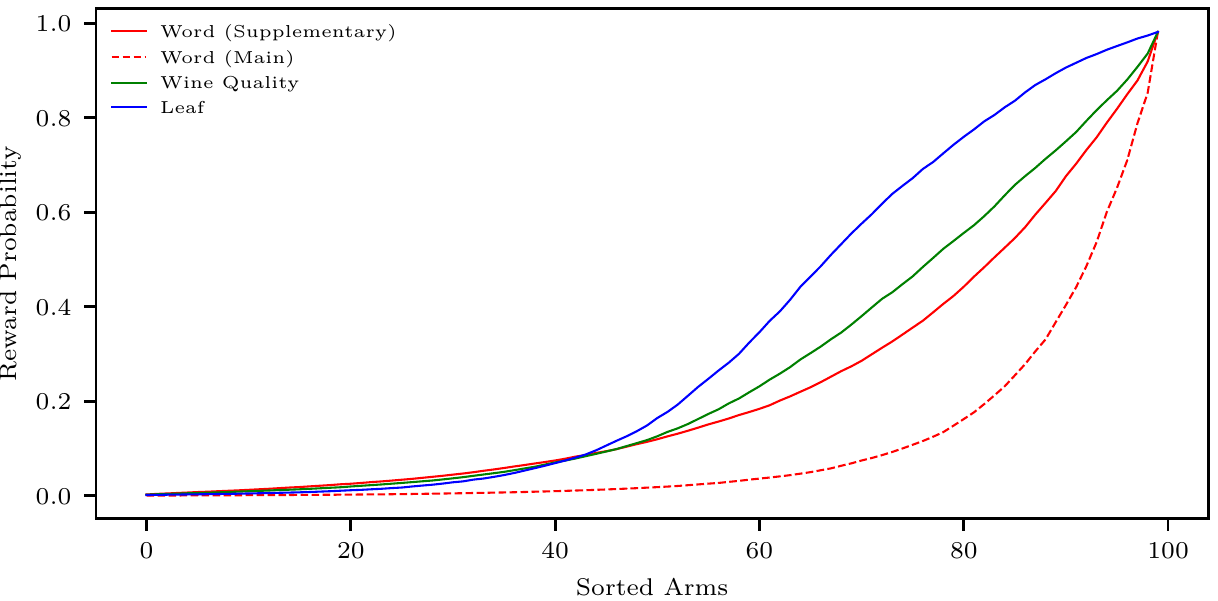}
		\centering
		\caption{The relevance profiles of the different datasets, generated by sorting the arms according to their reward probabilities and taking the mean over replications. For example, a point at (80,0.6) should be read as there are 20 arms with $\geq 0.6$ reward probability.}
		\label{fig:reward_landscape}
	\end{figure*}

\begin{figure*}[t]
		\begin{minipage}{.5\textwidth}
			\centering
			\includegraphics{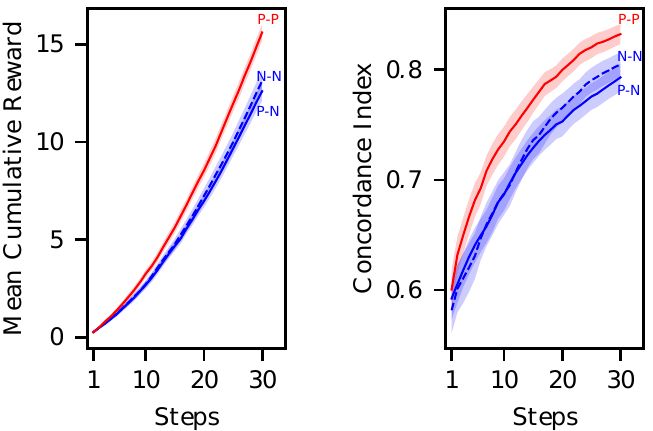}
		\end{minipage}%
		\begin{minipage}{.5\textwidth}
			\centering
			\includegraphics{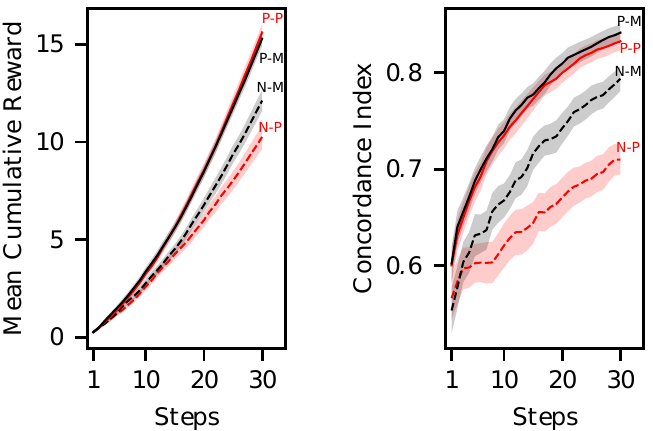}
		\end{minipage}
		\caption{Replication of the simulated experiment for the supplementary relevance profile setting in the Word dataset.}
		\label{fig:word_dense}
	\end{figure*}

\begin{figure*}[t]
		\begin{minipage}{.5\textwidth}
			\centering
			\includegraphics{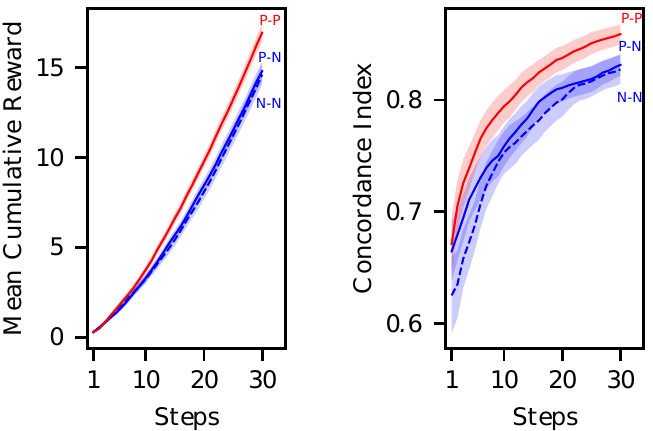}
		\end{minipage}%
		\begin{minipage}{.5\textwidth}
			\centering
			\includegraphics{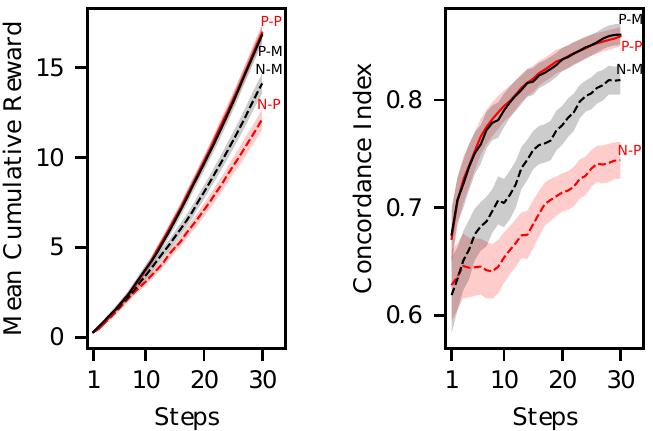}
		\end{minipage}
		\caption{Replication of the simulated experiment in the Wine Quality dataset.}
		\label{fig:wine_dense}
	\end{figure*}

\begin{figure*}[t]
		\begin{minipage}{.5\textwidth}
			\centering
			\includegraphics{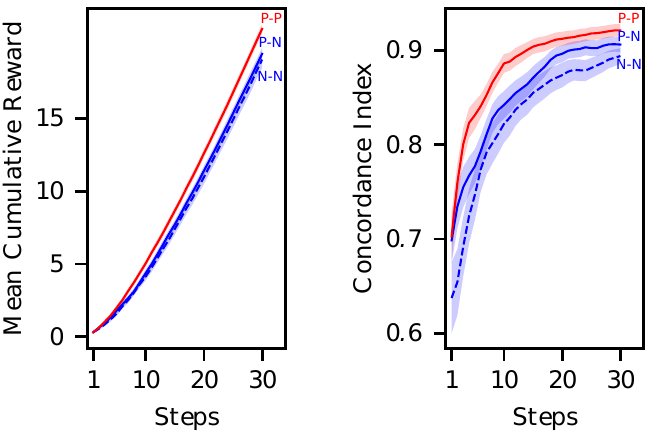}
		\end{minipage}%
		\begin{minipage}{.5\textwidth}
			\centering
			\includegraphics{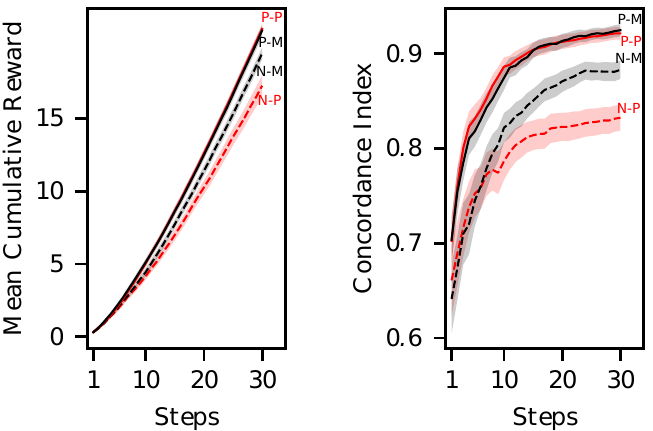}
		\end{minipage}
		\caption{Replication of the simulated experiment in the Leaf dataset.}
		\label{fig:leaf_dense}
	\end{figure*}

\begin{figure*}[h!]
		\includegraphics{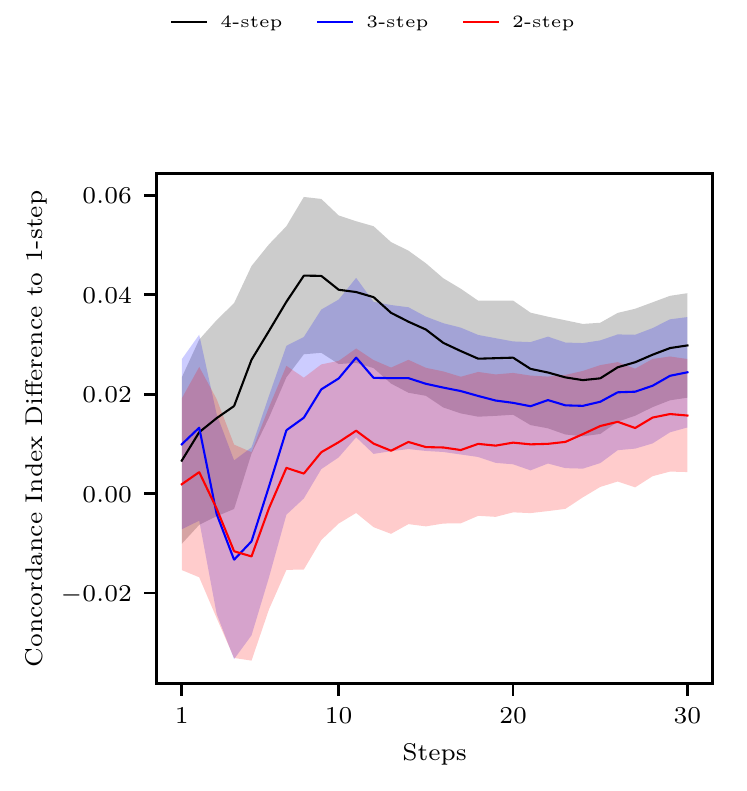}
		\centering
		\caption{
			Concordance index results for the multi-step experiment from the main text.}
		\label{fig:mla_sparse}
	\end{figure*}

\begin{figure*}[t]
		\begin{minipage}{.5\textwidth}
			\centering
			\includegraphics{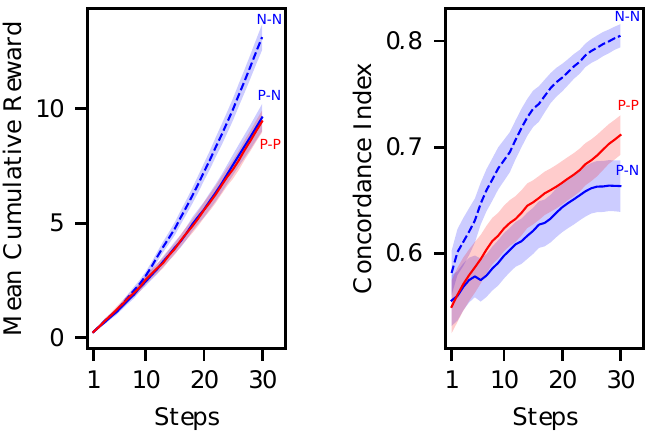}
		\end{minipage}%
		\begin{minipage}{.5\textwidth}
			\centering
			\includegraphics{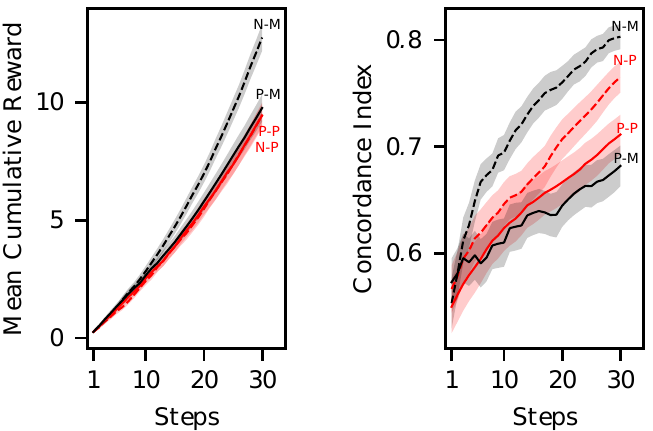}
		\end{minipage}
		\caption{Replication of the simulated experiment for the supplementary setting in the Word dataset with teacher's optimality parameter $\beta=5$.}
		\label{fig:word_beta5}
	\end{figure*}

\begin{figure*}[t]
		\begin{minipage}{.5\textwidth}
			\centering
			\includegraphics{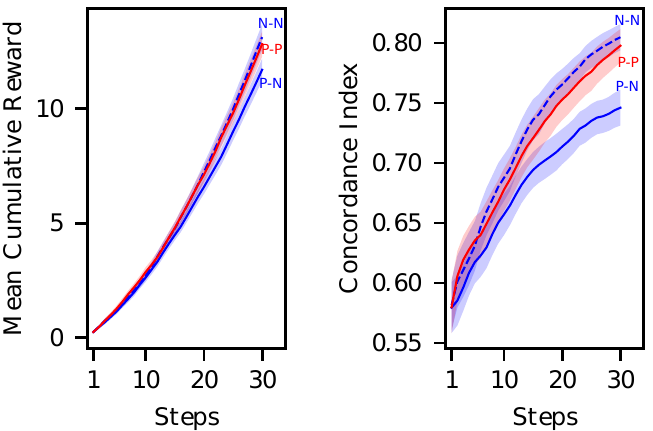}
		\end{minipage}%
		\begin{minipage}{.5\textwidth}
			\centering
			\includegraphics{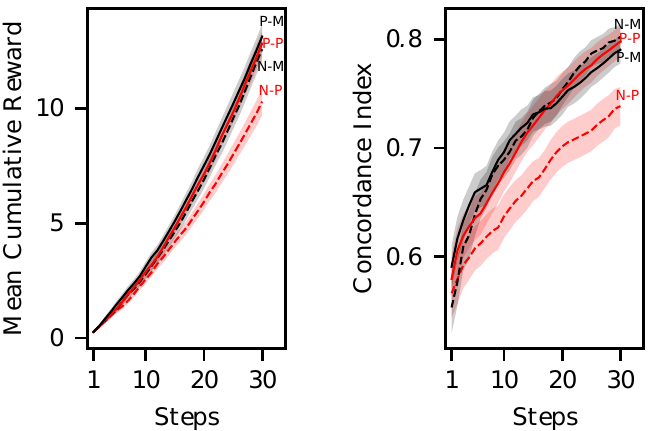}
		\end{minipage}
		\caption{Replication of the simulated experiment for the supplementary setting in the Word dataset with teacher's optimality parameter $\beta=10$.}
		\label{fig:word_beta10}
	\end{figure*}

\begin{figure*}[t]
		\begin{minipage}{.5\textwidth}
			\centering
			\includegraphics{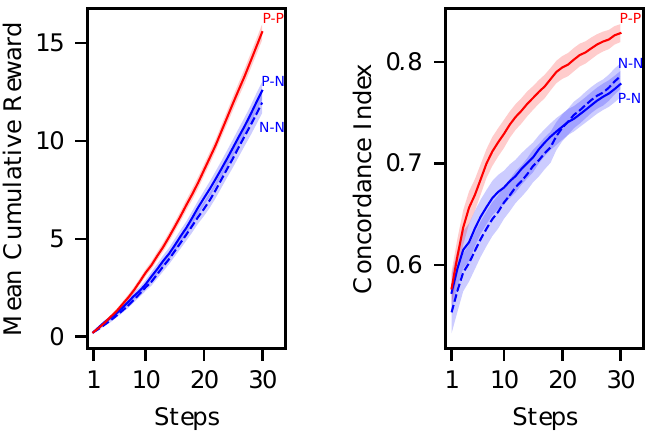}
		\end{minipage}%
		\begin{minipage}{.5\textwidth}
			\centering
			\includegraphics{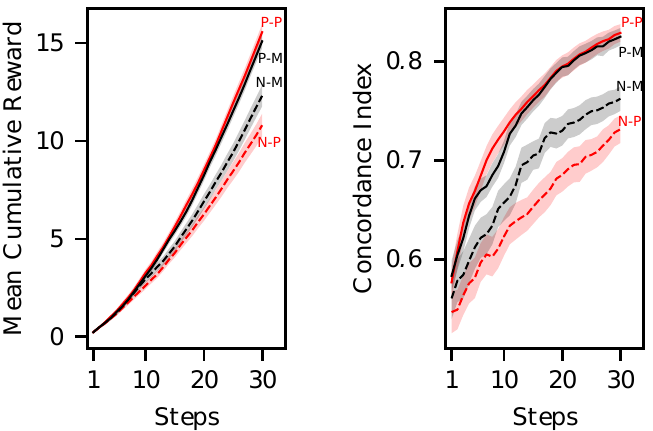}
		\end{minipage}
		\caption{Replication of the simulated experiment for the supplementary setting with 500 arms in the Word dataset.}
		\label{fig:word_500}
	\end{figure*}

	\begin{itemize}
		\item Figure~\ref{fig:reward_landscape}: The relevance profiles of the different datasets demonstrating the density of the reward probabilities. Word (Main) is the setting used in the results of the main text.
		\item Figure~\ref{fig:word_dense}: Replication of the simulated experiment in the Word dataset using the supplementary setting.
		\item Figure~\ref{fig:wine_dense}: Replication of the simulated experiment in the Wine Quality dataset.
		\item Figure~\ref{fig:leaf_dense}: Replication of the simulated experiment in the Leaf dataset.
		\item Figure~\ref{fig:mla_sparse}: Concordance index results for the multi-step experiment from the main text.
		\item Figure~\ref{fig:word_beta5}: Replication of the simulated experiment in the Word dataset using the supplementary setting, with teacher's optimality parameter $\hat{\beta}=5$ and teacher model parameter $\beta = 5$. This demonstrates that highly suboptimal teaching degrades learning performance. 
		\item Figure~\ref{fig:word_beta10}: Replication of the simulated experiment in the Word dataset using the supplementary setting, with teacher's optimality parameter $\hat{\beta}=10$ and teacher model parameter $\beta = 10$.
		\item Figure~\ref{fig:word_500}: Replication of the simulated experiment with 500 arms in the Word dataset using the supplementary setting.
	\end{itemize}

	\clearpage

	\section{User study details}
	
\begin{figure*}[h!]
		\includegraphics[width=1\columnwidth]{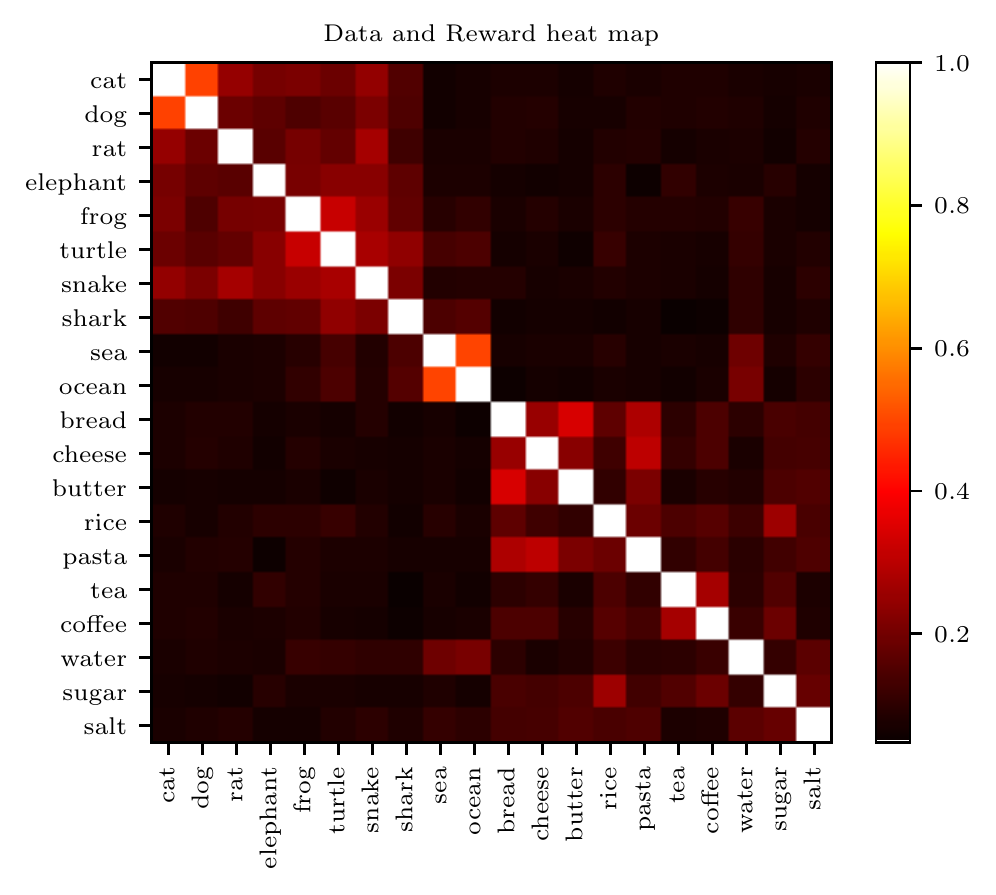}
		\centering
		\caption{
			User study data and ground truth rewards. The matrix represents the feature vectors of each word considered in the user study. The ground truth reward values for each target word are represented by the values in the corresponding row.}
		\label{fig:Data_and_rewards}
	\end{figure*}

	We conducted a proof-of-concept user study using a subset of 20 words (from the Word dataset) on ten university students and researchers (3 females and 7 males). The goal of the study was introduced to the participants as helping a system to find a target word, as fast as possible, by sequentially providing binary answers (yes/no) to the system's questions (15 question budget) about the relevance of different words to the target word. The target word was given to the participants at the beginning of each round. Since only 20 words were selected for the user study, we skipped the PCA pre-processing step (considering it is hard to detect information from noise when the number of data is much smaller than the dimension) and instead used a pairwise radial basis function kernel between the words in the original 300 dimension to reduce the dimension to 20. Furthermore, to reduce the variance in the sequence of questions in both models, we used Bayes-UCB \cite{kaufmann2012bayesian} instead of Thompson sampling as the system's outer-most arm selection strategy. The list of considered words along with their feature vectors is shown in Figure~\ref{fig:Data_and_rewards}. The resulting data matrix was also used as the ground truth reward function for each target word (asking about the target word gains reward one and others noticeably less than one).
	
	The study was repeated for naive and mixture models (model means learner's model of the teacher) in randomised order and for twenty rounds (each word chosen once as the target word). The planning teacher model in the mixture had one-step planning horizon ($T=1$). The starting question of each round was chosen randomly for each user and target but it was the same between the two models. Two practice rounds (one with each model) were completed in the beginning of the study. The user interface and time delays between questions were identical between the two models and the participants were naive about which system they were interacting with. The questions were in the form of "Is \textit{word} Relevant?" and the participants could answer by typing "y" (Yes) or "n" (No) in the terminal and pressing enter. Each round would end after 15 questions and answers. The users were not under any time pressure and the study took on average 75 minutes. 
	
	The participants were compensated by a movie ticket upon completion of the study. The task performance was incentivised by providing an extra movie ticket if the participant was able to help the model find the target word in fewer steps than a certain threshold. All participants signed a standard consent form prior to the study. All user studies were performed on a Windows 10 laptop.
	
	\begin{figure*}[h!]
		\includegraphics[width=1\columnwidth]{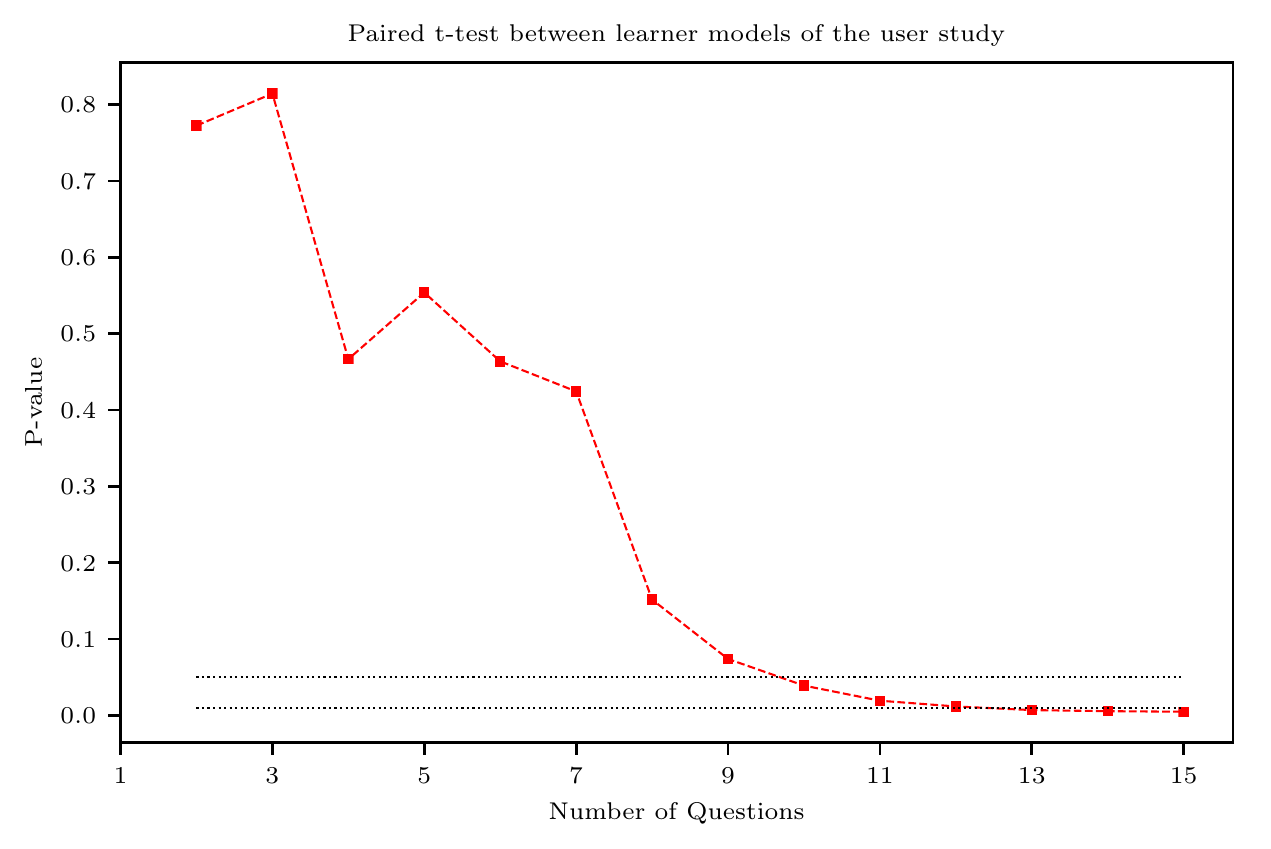}
		\centering
		\caption{
			P-value for paired sample t-test between average cumulative reward of the mixture and naive models of ten participants of the user study at each iteration. The black dashed lines show the 0.05 and 0.01 thresholds. The mixture model achieved significantly higher cumulative reward after 12 questions.}
		\label{fig:p_values_user_study}
	\end{figure*}
	
	\clearpage
	
	\renewcommand\refname{Supplementary References}

	%\bibliography{neurips_references}
	\bibliographystyle{unsrtnat}

\end{document}